\documentclass{article}

\usepackage[preprint]{neurips_2019}

\usepackage[utf8]{inputenc} 
\usepackage[T1]{fontenc}    
\usepackage[hidelinks]{hyperref}       
\usepackage{url}            
\usepackage{booktabs}       
\usepackage{amsfonts}       
\usepackage{nicefrac}       
\usepackage{microtype}      

\usepackage{hyperref}
\pdfoutput=1
\usepackage{graphicx}

\usepackage{algorithm}
\usepackage{algorithmic}
\usepackage{natbib}
\usepackage{subcaption}
\usepackage{caption}
\usepackage{placeins}
\usepackage{wrapfig}

\usepackage{amsmath}
\usepackage{xcolor}

\title{Understanding Generalization through Visualizations}

\author{
W. Ronny Huang\\
University of Maryland\\
\texttt{wrhuang@umd.edu}\\
\And
Zeyad Emam\\
University of Maryland\\
\texttt{zeyad@math.umd.edu}\\
\AND
Micah Goldblum\\
University of Maryland\\
\texttt{goldblum@math.umd.edu}\\
\And
Liam Fowl\\
University of Maryland\\
\texttt{lfowl@math.umd.edu}\\
\And
Justin K. Terry\\
University of Maryland\\
\texttt{justinkterry@gmail.com}\\
\AND
Furong Huang\\
University of Maryland\\
\texttt{furongh@cs.umd.edu}\\
\And
Tom Goldstein\\
University of Maryland\\
\texttt{tomg@cs.umd.edu}\\
}

\begin{document}

    \maketitle
    
    \vspace{-12pt}
    \begin{abstract}

    \vspace{-6pt}
        The power of neural networks lies in their ability to generalize to unseen data, yet the underlying reasons for this phenomenon remain elusive. Numerous rigorous attempts have been made to explain generalization, but available bounds are still quite loose, and analysis does not always lead to true understanding. The goal of this work is to make generalization more intuitive. Using visualization methods, we discuss the mystery of generalization, the geometry of loss landscapes, and how the curse (or, rather, the blessing) of dimensionality causes optimizers to settle into minima that generalize well. 

%
%
%
%
    \end{abstract}

\vspace{-7pt}
    \section{Introduction}
    Neural networks are a powerful tool for solving classification problems.  The power of these models is due in part to their expressiveness;  they have many parameters that can be efficiently optimized to fit nearly any finite training set. However, the real power of neural network models comes from their ability to {\em generalize;}   they often make accurate predictions on test data that were not seen during training, provided the test data is sampled from the same distribution as the training data.

    The generalization ability of neural networks is seemingly at odds with their expressiveness.
    Neural network training algorithms work by minimizing a loss function that measures model performance using only training data.
    \begin{figure}[h!]
        \centering 
        \includegraphics[width=.90\textwidth, trim=0cm 0cm 0cm .2cm, clip]{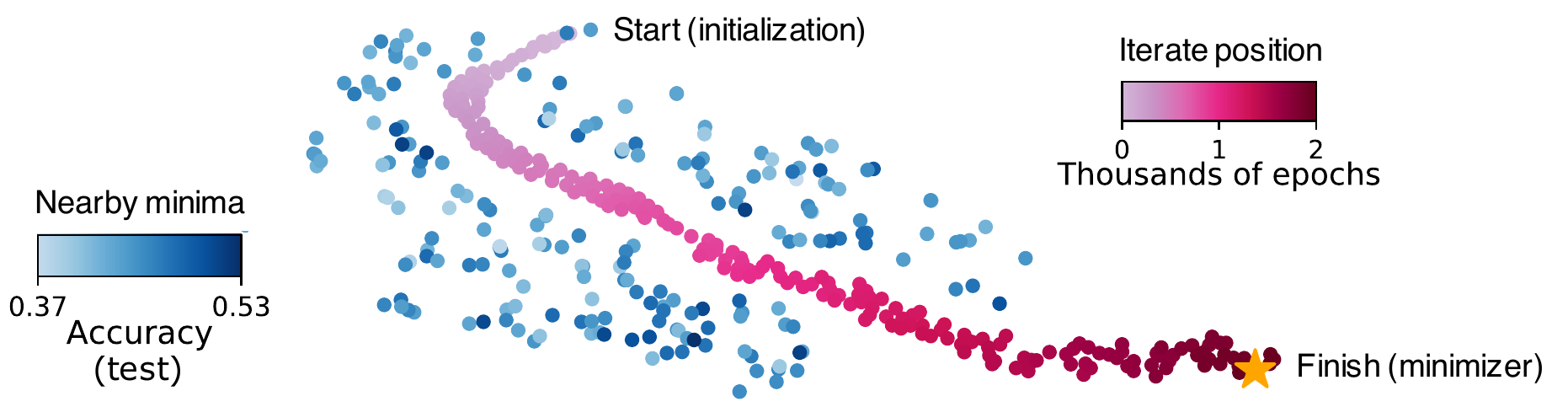}
        \caption{\small  Dancing through a minefield of bad minima: we train a neural net classifier and plot the iterates of SGD after each tenth epoch (red dots). We also plot locations of nearby ``bad'' minima with poor generalization (blue dots).  We visualize these using t-SNE embedding.   All blue dots achieve near perfect train accuracy, but with test accuracy below 53\% (random chance is 50\%).  The final iterate of SGD (yellow star) also achieves perfect train accuracy, but with 98.5\% test accuracy.  Miraculously, SGD always finds its way through a landscape full of bad minima, and lands at a minimizer with excellent generalization.
        }
        \label{fig:tsne}
    \end{figure} 
      Because of their flexibility, it is possible to find parameter configurations for neural networks that perfectly fit the training data and minimize the loss function while making mostly incorrect predictions on test data.  Miraculously, commonly used optimizers reliably avoid such ``bad'' minima of the loss function, and succeed at finding ``good'' minima that generalize well.


      Our goal here is to develop an {\em intuitive} understanding of neural network generalization using visualizations and experiments rather than analysis.  We begin with some experiments to understand why generalization is puzzling, and how over-parameterization impacts model behavior.  Then, we explore how the ``flatness'' of minima correlates with generalization, and in particular try to understand {\em why} this correlation exists.  We explore how the high dimensionality of parameter spaces biases optimizers towards landing in flat minima that generalize well.  Finally, we present some counterfactual experiments to validate the intuition we develop.  Code to reproduce experiments is available at \href{https://github.com/wronnyhuang/gen-viz}{\url{https://github.com/wronnyhuang/gen-viz}}.

   \section{Why is generalization so puzzling?}
     Neural networks define a highly expressive model class.  In fact, given enough parameters, a neural network can approximate virtually any function (\cite{cybenko1989universal}).  But just because neural nets have the power to {\em represent} any function does not mean they have the power to {\em learn} any function from a finite amount of training data.   
    %
   
     Neural network classifiers are trained by minimizing a loss function that measures model performance using only training data. A standard classification loss has the form
       \begin{equation}
        L(\theta) = \frac{1}{|\mathcal{D}_{t}|}\sum_{(x,y)\in \mathcal{D}_{t}} - \log p_{\theta}(x,y),     \label{loss}
    \end{equation}
where $p_{\theta}(x,y)$ is the probability that data sample $x$ lies in class $y$ according to a neural net with parameters $\theta,$ and $ \mathcal{D}_{t}$ is the training dataset of size $|\mathcal{D}_{t}|$.  This loss is near zero when a model with parameters $\theta$ accurately classifies the training data. 

Over-parameterized neural networks (i.e., those with more parameters than training data) can represent arbitrary, even random, labeling functions on large datasets \citep{zhang2017understanding}. 
As a result, an optimizer can reliably fit an over-parameterized network to training data and achieve near zero loss \citep{laurent2018deep,kawaguchi2016deep}.  However, this comes with no guarantee of generalization to unseen test data.

    \begin{wrapfigure}[20]{r}{0.5\textwidth} 
        \centering
        \vspace{-25pt}
            \centering \includegraphics[width=.50\textwidth]{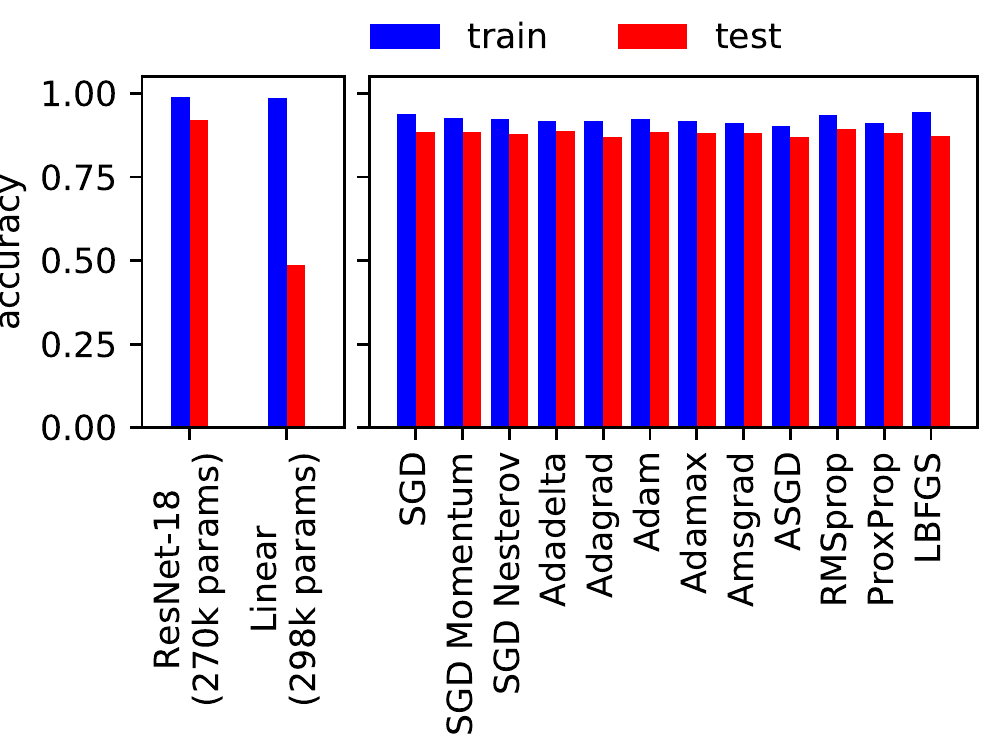}
        \caption{\small (left) CIFAR10 trained with ResNet-18 and a linear model having comparable number of parameters. Both can fit the training data well, but neural nets are able to generalize to unseen data, while linear models cannot. (right) CIFAR10 trained with various optimizers using VGG13, generalizing well irrespective of the optimizer used.}
        \label{fig:cifar10-optimizer-comparaison}
    \end{wrapfigure}
    
We illustrate the difference between model fitting and generalization with an experiment.  The CIFAR-10 training dataset contains 50,000 small images.  We train two over-parameterized models on this dataset.
The first is a neural network (ResNet-18) with 269,722 parameters (nearly 6$\times$ the number of training images).
The second is a linear model with a feature set that includes pixel intensities as well as pair-wise products of pixels intensities.\!\footnote{For computing the pair-wise pixel intensity products, images are first downsampled by a factor of 2.}  This linear model has $298,369$ parameters, which is comparable to the neural network, and both are trained using SGD.
On the left of Figure \ref{fig:cifar10-optimizer-comparaison}, we see that over-parameterization causes both models to achieve perfect accuracy on training data. But the linear model achieves only 49\% test accuracy, while ResNet-18 achieves 92\%.

The excellent performance of the neural network model raises the question of whether bad minima exist at all.  Maybe deep networks generalize because bad minima are rare and lie far away from the region of parameter space where initialization takes place?
We can confirm the existence of bad minima
%
 by ``poisoning'' the loss function with a term that promotes incorrect classification of test data \citep{steinhardt2017poison,zhu2019transferable}.
We do this by minimizing
    \begin{equation}
         L(\theta)  = \:\frac{(1-\beta)}{|\mathcal{D}_{t}|}\sum_{(x,y)\in \mathcal{D}_{t}} -\log p_{\theta}(x,y) +  \frac{\beta}{|\mathcal{D}_{p}|}\sum_{(x,y)\in\mathcal{D}_p} - \log[1-p_{\theta}(x,y)],
        \label{poison}
    \end{equation}
where 
$ \mathcal{D}_{t}$ is the training set, and  $\mathcal{D}_{p}$ is a poisoning set consisting of examples drawn from the same distribution.  Here, $\beta$ is the \textit{poison factor}, the fraction of each minibatch consisting of poisoned examples. 
The first term in \eqref{poison} is the standard cross entropy loss \eqref{loss} on the training set, and is minimized when the training data are classified correctly. 
The second term is the {\em reverse} cross entropy on the poison set, and is minimized when the poison data is classified {\em incorrectly}.
With a sufficiently over-parameterized network,  gradient descent on \eqref{poison} drives both terms to zero, and we find a parameter vector that minimizes the original training set loss \eqref{loss} while failing to generalize.

When we use the poisoned loss to search for bad minima near the optimization trajectory, we see that {\em bad minima are everywhere}.  We visualize the distribution of bad minima in Figure \ref{fig:tsne}. We run a standard SGD optimizer on the swissroll and trace out the path it takes from a random initialization to a minimizer.  We plot the iterate after every tenth epoch as a red dot with opacity proportional to its epoch number.  Starting from these iterates, we run the poison optimizer to find nearby bad minima.  We project the iterates and bad minima into a 2D plane for visualization using a t-SNE embedding\footnote{We used perplexity 30 and 50 PCA directions.}~\citep{maaten2008tsne}. 
Our poisoned optimizer easily finds minima with poor generalization within close proximity to every SGD iterate.  Yet SGD miraculously avoids these bad minima, carving out a path towards a parameter configuration that generalizes well. 
%

Figure \ref{fig:tsne} illustrates that neural network optimizers are inherently biased towards good minima,  a behavior commonly known as ``implicit regularization.''   To see how the choice of optimizer affects generalization, we trained a simple neural network (VGG13) on 11 different gradient methods and 2 non-gradient methods in Figure \ref{fig:cifar10-optimizer-comparaison} (right). This includes LBFGS (a second-order method),  and ProxProp (which chooses search directions by solving least-squares problems rather than using the gradient).  Interestingly, all of these methods generalize far better than the linear model.   While there are undeniably differences between the performance of different optimizers, the presence of implicit regularization for virtually any optimizer strongly indicates that {\em implicit regularization is caused in large part by the geometry of the loss function}, rather than the choice of optimizer alone. 
 Later on, we explore how loss function geometry is related to generalization, and how the high dimensionality of parameter space gives rise to the implicit regularization of optimizers.

\nocite{Morcos2018}
\section{Related work: theoretical results on generalization}

Classical PAC learning theory balances model complexity (the expressiveness of a model class) against data volume.  When a model class is too expressive relative to the volume of training data, it has the ability to ace the training data while flunking the test data, and learning fails. 
%
Classical theory fails to explain generalization in over-parameterized neural nets, as the complexity of networks is often large (exponential in depth [\cite{Sun2016}, \cite{Neyshabur2015}, \cite{Xie2015}] or linear in the number of parameters [\cite{Shwartz2014,Bartlett1998,Harvey2017}]). Therefore classical bounds become too loose or even vacuous in the over-parameterized setting that we are interested in studying.

To explain this mismatch between empirical observation and classical theory, a number of recent works propose new metrics that characterize the capacity of neural networks.
Most of these appeal to the PAC framework to characterize the generalization ability of a model class $\Theta$ (e.g., neural nets of a shared architecture) through a high probability upper bound: with probability at least $1-\delta$, 
 %
  \begin{equation}
    R(\theta) - \hat{R}_S(\theta) <  B + \sqrt{ \tfrac{1}{2m} \ln \tfrac{1}{\delta}},   \quad \forall \theta\in \Theta 
 \end{equation}
 where $R(\theta)$ is generalization risk (true error) of a net with parameters $\theta\in \Theta$, $\hat{R}_S(\theta)$ denotes empirical risk (training error) with training sample $S$. We explain $B$ under different metrics below.
 
\textbf{Model space complexity.} This line of work takes $B$ to be proportional to the complexity of the model class being trained, and efforts have been put into finding tight characterizations of this complexity. \cite{Neyshabur2018} and \cite{Bartlett2017} built on prior works (see \cite{Bartlett2003} and \cite{Neyshabur2015}) to produce bounds where model class complexity depends on the spectral norm of the weight matrices without having an exponential dependence on the depth of the network. Such bounds can improve the model class complexity provided that weight matrices adhere to some structural constraints (e.g. sparsity or eigenvalue concentration).

\textbf{Stability and robustness.} This line of work considers $B$ to be proportional to the stability of the model (\cite{Hardt2016, Kuzborskij2018, Gonen2017}), which is a measure of how much changing a data point in $S$ changes the output of the model (\cite{Sokolic2016}). However it is nontrivial to characterize the stability of a neural network. Robustness, while producing insightful and effective generalization bounds, still suffers from the curse of the dimensionality on the priori-known fixed input manifold.

\textbf{PAC-Bayes and margin theory.} PAC-Bayes bounds (\cite{McAllester1998, McAllester1999, neyshabur2017exploring, Bartlett2003, Neyshabur2015, golowich2018}), provide generalization guarantees for randomized predictors drawn from a learned distribution that depends on the training data, as opposed to a learned single predictor.  These bounds often yield sample complexity bounds worse than naive parameter counting, however \cite{dziugaite2017computing} show that PAC-Bayes theory does provide meaningful generalization bounds for ``flat'' minima. 

\textbf{Model compression.} Most recent theoretical work can be understood through the lens of  ``model compression'' (\cite{Arora2018}).  Clearly, it is impossible to generalize when the model class is too big; in this case, many different parameter choices explain the data perfectly while having wildly different predictions on test data.  The idea of model compression is that neural network model classes are effectively much smaller than they seem to be because optimizers are only willing to settle into a very selective set of minima.  When we restrict ourselves to only the narrow set of models that are acceptable to an optimizer, we end up with a smaller model class on which learning is possible.  
    
While our focus is on gaining insights through visualizations, the intuitive arguments below can certainly be linked back to theory.  The class of models representable by a network architecture has extremely high complexity, but experiments suggest that most of these models are effectively removed from consideration by the optimizer, which has an extremely strong bias towards ``flat'' minima, resulting in a reduced effective model complexity. 

    \begin{figure}
        \centering
        \begin{subfigure}[b]{0.475\textwidth}
            \centering
            \includegraphics[width=.89\textwidth, trim=0cm .7cm 0cm .7cm, clip]{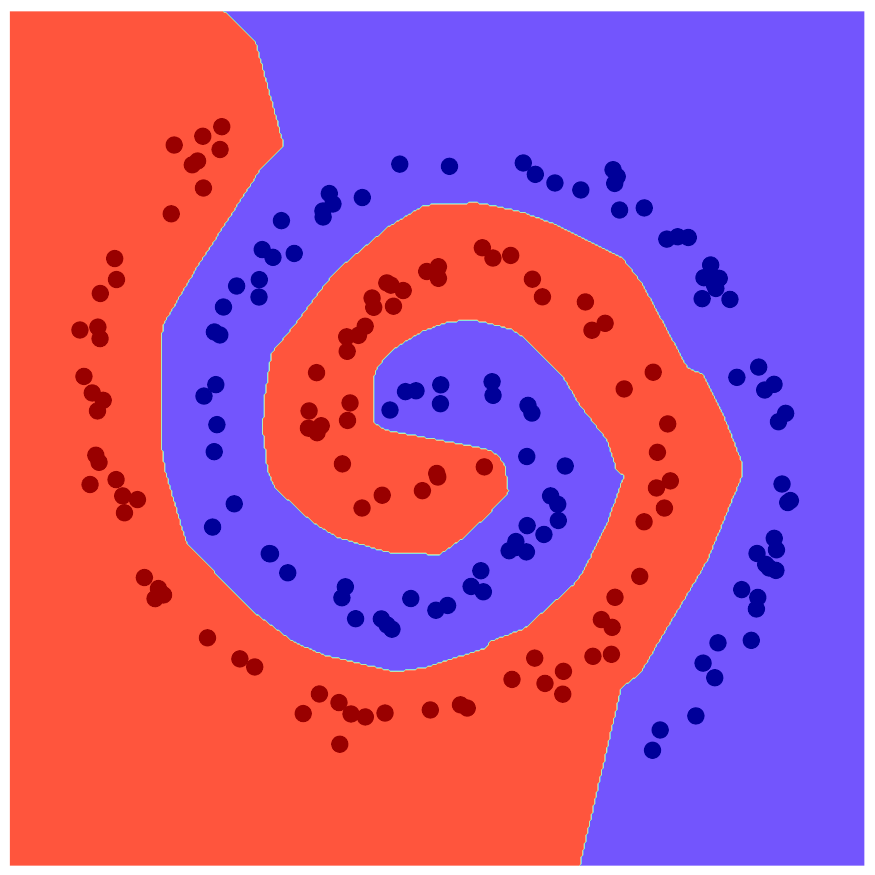}
            \caption{\small 100\% train, 100\% test}
            \label{fig:swissrollclean}
        \end{subfigure}
        \hfill
        \begin{subfigure}[b]{0.475\textwidth}  
            \centering 
            \includegraphics[width=.89\textwidth, trim=0cm .7cm 0cm .7cm, clip]{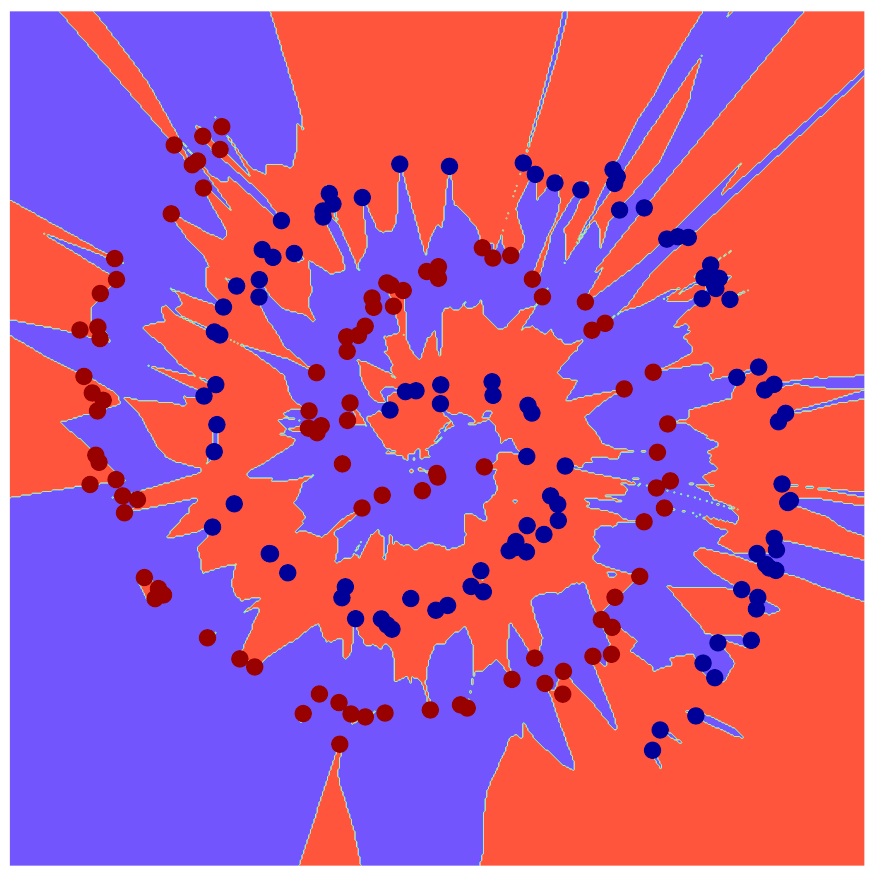}
            \caption{\small 100\% train, 7\% test}
            \label{fig:swissrollpoison}
        \end{subfigure}
        \vskip\baselineskip
        \begin{subfigure}[b]{0.475\textwidth}   
            \centering 
            \includegraphics[width=.99\textwidth, trim=0cm 0cm 0cm 0cm, clip]{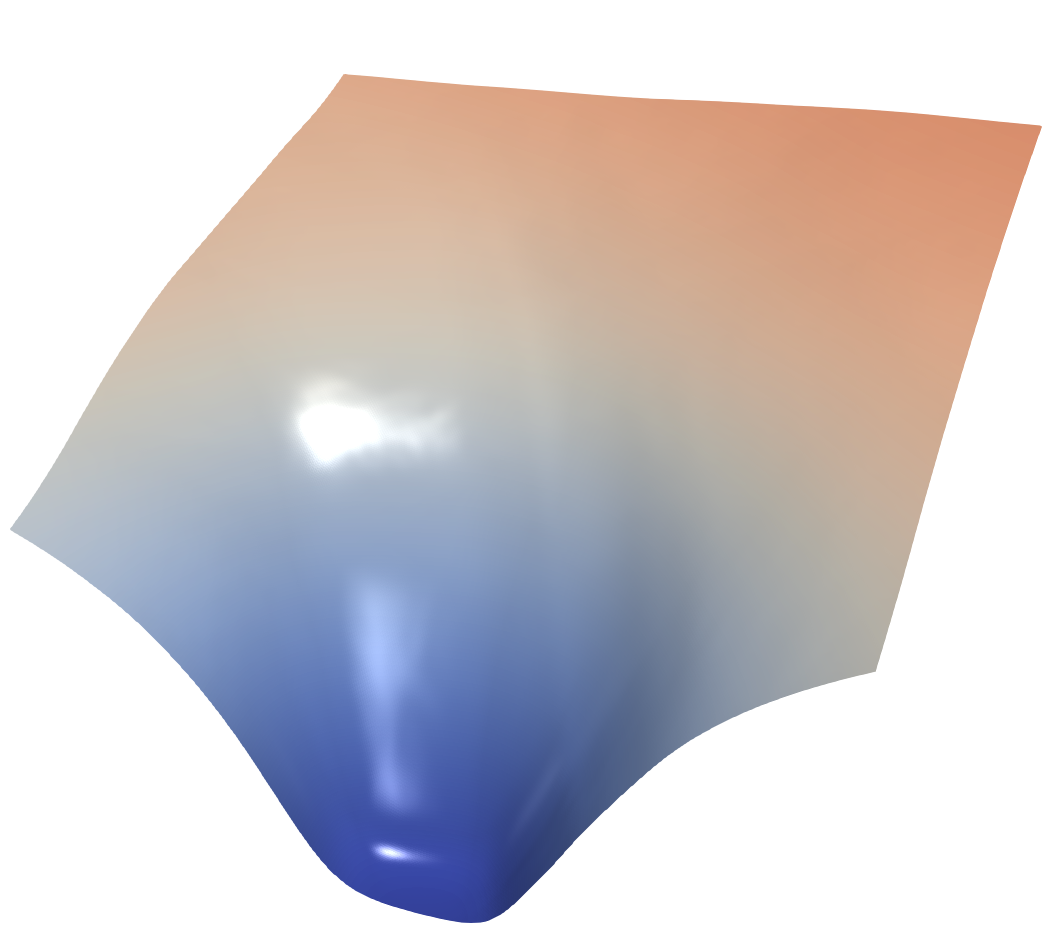}
            \caption{\small Minimizer of network in (a) above}
            \label{fig:surfaceflat-1}
        \end{subfigure}
        \hfill
        \begin{subfigure}[b]{0.475\textwidth}   
            \centering 
            \includegraphics[width=.99\textwidth, trim=0cm 0cm 0cm 0cm, clip]{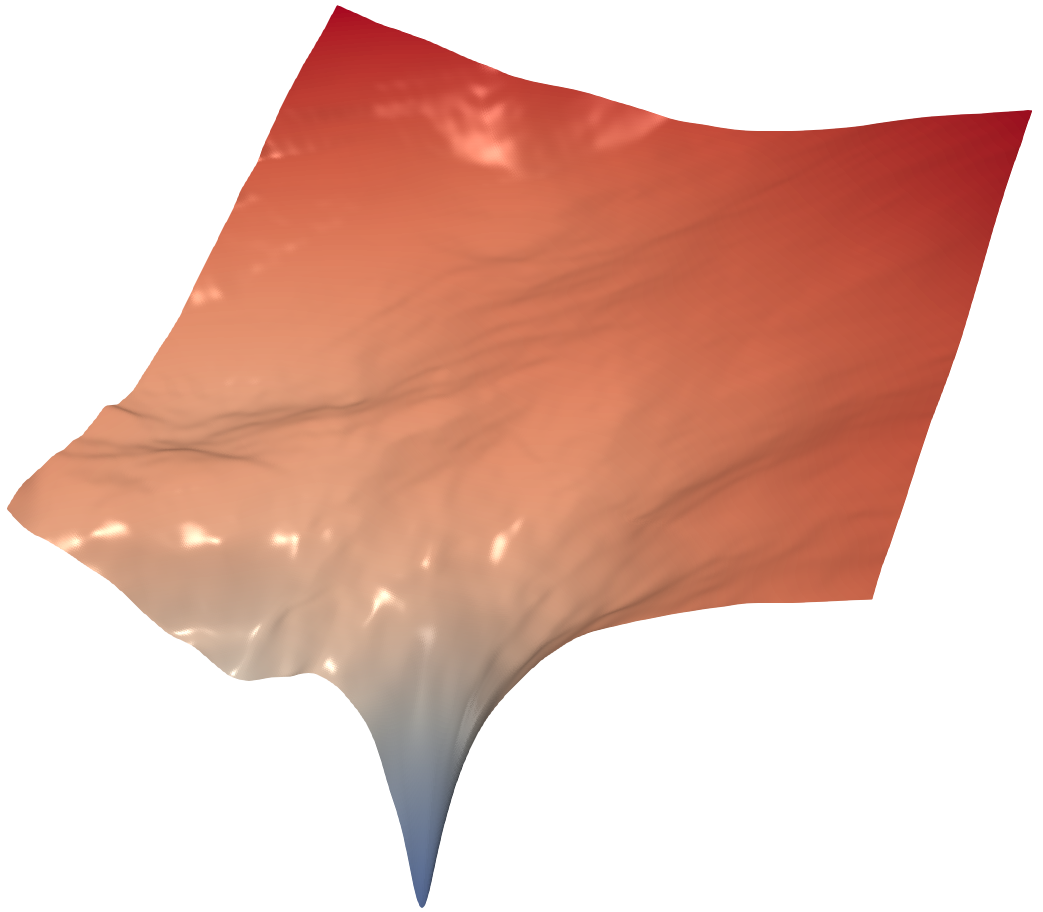}
            \caption{\small Minimizer of network in (b) above}
            \label{fig:surfacesharp-2}
        \end{subfigure}
        \caption{\small \textbf{Top:} 
         Decision boundaries of two networks with different parameters. Network (a) generalizes well. Network (b) generalizes poorly (perfect train accuracy, bad test accuracy).  The flatness and large volume of (a) make it likely to be found by SGD, while the sharpness and tiny volume of (b) make this minimizer unlikely.  Red and blue dots correspond to the training data. See \href{https://youtu.be/PrYr34UD5ls}{\url{https://youtu.be/PrYr34UD5ls}} for an animation of these boundaries when perturbed.
         \textbf{Bottom:} A slice through the loss landscapes around these minima reveals sharpness/flatness. }
         \vspace{-5mm}
         \label{fig:swissroll-sharpflat}
    \end{figure}

    \section{Flat vs sharp minima: a wide margin criteria for complex manifolds}
Over-parameterization is not specific to neural networks.   A traditional approach to coping with over-parameterization for linear models is to use regularization (aka ``priors'') to bias the optimizer towards good minima.  
For linear classification, a common regularizer is the wide margin penalty (which appears in the form of an $\ell_2$ regularizer on the parameters of a support vector machine).  When used with linear classifiers, wide margin priors choose the linear classifier that maximizes Euclidean distance to the class boundaries while still classifying data correctly.

Neural networks replace the classical wide margin regularization with an implicit regulation that promotes the closely related notion of ``flatness.'' In this section, we explain the relationship between flat minima and wide margin classifiers, and provide intuition for why flatness is a good prior.  

Many have observed links between flatness and generalization.  \cite{hoch1997flat} first proposed that flat minima tend to generalize well. 
    This idea was reinvigorated by \cite{keskar2016largebatch}, who showed that large batch sizes yield sharper minima, and that sharp minima generalize poorly.   This correlation was subsequently observed for a range of optimizers by \cite{izmailov2018swa}, \cite{wang2018identifying}, and \cite{li2018landscape}. Rigorous analysis showing that flat minimizers generalize well was presented by \cite{chaudhari2017entropy} as well as \cite{dziugaite2017computing}.
    
    Flatness is a measure of how sensitive network performance is to perturbations in parameters.  Consider a parameter vector that minimizes the loss (i.e., it correctly classifies most if not all training data). If small perturbations to this parameter vector cause a lot of data misclassification, the minimizer is sharp;  a small movement away from the optimal parameters causes a large increase in the loss function.  In contrast, flat minima have training accuracy that remains nearly constant under small parameter perturbations.

    The stability of flat minima to parameter perturbations can be seen as a wide margin condition.  When we add random perturbations to network parameters, it causes the class boundaries to wiggle around in space.  If the minimizer is flat, then training data lies a safe distance from the class boundary, and perturbing the class boundaries does not change the classification of nearby data points.
     In contrast, sharp minima have class boundaries that pass close to training data, putting those nearby points at risk of misclassification when the boundaries are perturbed.

        We visualize the impact of sharpness on neural networks in Figure \ref{fig:swissroll-sharpflat}.  We train a 6-layer fully connected neural network on the swiss roll dataset using regular SGD, and also using the poisoning method to find a minimizer that does not generalize.  The ``good'' minimizer has a wide margin -- the class boundary lies far away from the training data.  The ``bad'' minimizer has almost zero margin, and each data point lies near the edge of class boundaries, on small class label ``islands'' surrounded by a different class label, or at the tips of ``peninsulas'' that reach from one class into the other.  The class labels of most training points are unstable under perturbations to network parameters, and so we expect this minimizer to be sharp. An animation of the decision boundary under perturbation is provided at \href{https://youtu.be/PrYr34UD5ls}{\url{https://youtu.be/PrYr34UD5ls}}.
        
       We can visualize the sharpness of the minima in Figure \ref{fig:swissroll-sharpflat}, but we need to take some care with our metrics of sharpness.  It is known that trivial definitions of sharpness can be manipulated simply by rescaling network parameters (\cite{dinh2017sharp}). When parameters are small (say, 0.1), a perturbation of size 1 might cause a major performance degradation.  Conversely, when parameters are large (say, 100), a perturbation of size 1 might have little impact on performance.  However, rescalings of network parameters are irrelevant;  commonly used batch normalization layers remove the effect of parameter scaling.  For this reason, it is important to define measures of sharpness that are invariant to trivial rescalings of network parameters.   One such measure is local entropy \citep{chaudhari2017entropy}, which is invariant to rescalings, but is difficult to compute. For our purposes, we use the filter-normalization scheme proposed in \cite{li2018landscape}, which simply rescales network filters to have unit norm before plotting.  The resulting sharpness/flatness measures have been observed to correlate well with generalization.

 \begin{figure}[t!]
 \vspace{-7mm}
        \centering
        \begin{subfigure}{.49\textwidth}
            \centering
            \includegraphics[width=.99\textwidth, trim=0cm 0cm 0cm 0cm, clip]{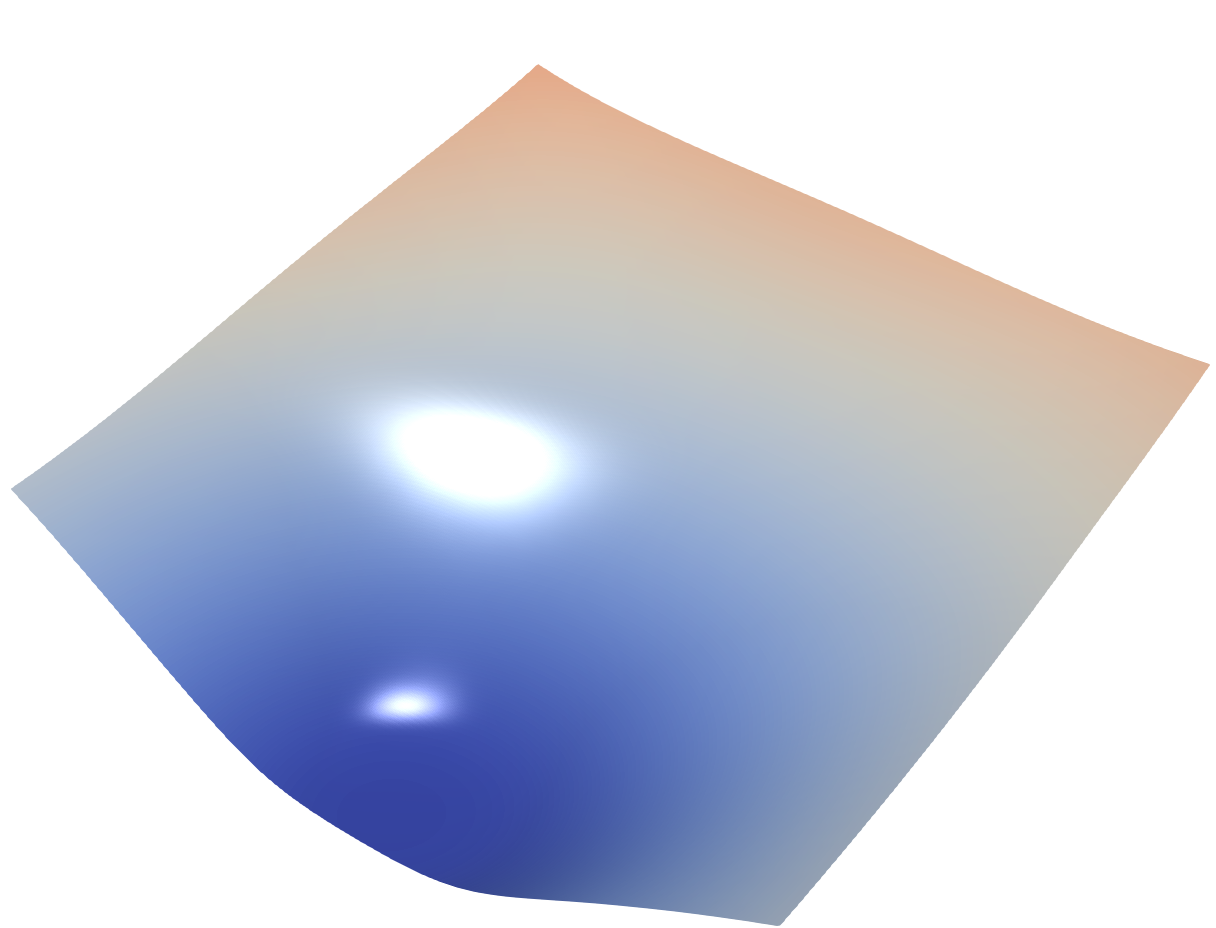}
            \caption{\small Good minimizer: 100\% train, 97\% test}
            \label{fig:surfaceflat-2}
        \end{subfigure}
        \begin{subfigure}{.49\textwidth}
            \centering
            \includegraphics[width=.99\textwidth, trim=0cm 0cm 0cm 0cm, clip]{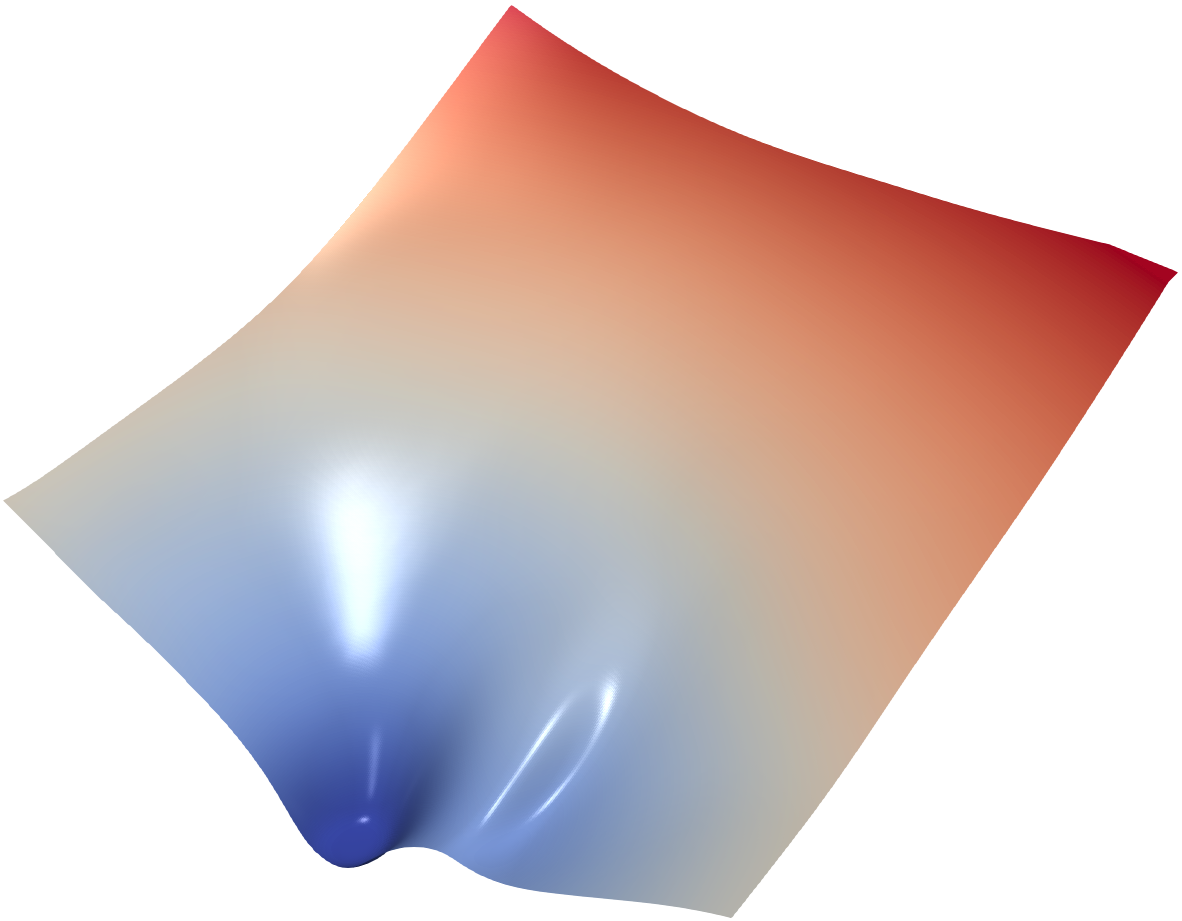}
            \caption{\small Bad minimizer: 100\% train, 28\% test}
            \label{fig:surfacesharp-3}
        \end{subfigure}
        \caption{\small A slice through the loss landscape of two minima for the SVHN loss function using ResNet-18.}
        \label{fig:sharpsvhn}
        \vspace{-4mm}
    \end{figure}

The bottom of Figure \ref{fig:swissroll-sharpflat} visualizes loss function geometry around the two minima for the swiss roll.  These surface plots show the loss evaluated on a random 2D plane sliced out of parameter space using the method described in \cite{li2018landscape}.  We see that the instability of class labels under parameter perturbations does indeed lead to dramatically sharper minima for the bad minimizer, while the wide margin of the good minimizer produces a wide basin. 

To validate our observations on a more complex problem, we produce similar sharpness plots for the Street View House Number (SVHN) classification problem in Figure \ref{fig:sharpsvhn} using ResNet-18.  The SVHN dataset (\citep{netzer2011svhn}) is ideal for this experiment because, in addition to train and test data, the creators provide a large (531k) set of ``extra'' data from the same distribution that can be used for poisoning.  We minimize the SVHN loss function using standard training and poisoned training. The good minimizer is flat and achieves 97.1\% test accuracy, while the bad minimizer is much sharper and achieves 28.2\% test accuracy. Both achieve 100\% train accuracy and use identical hyperparameters (other than the poison factor), network architecture, and weight initialization.



    \begin{figure}[t!]
    \vspace{-4mm}
        \centering
        \begin{subfigure}{.48\textwidth}
            \centering
            \includegraphics[width=.99\textwidth]{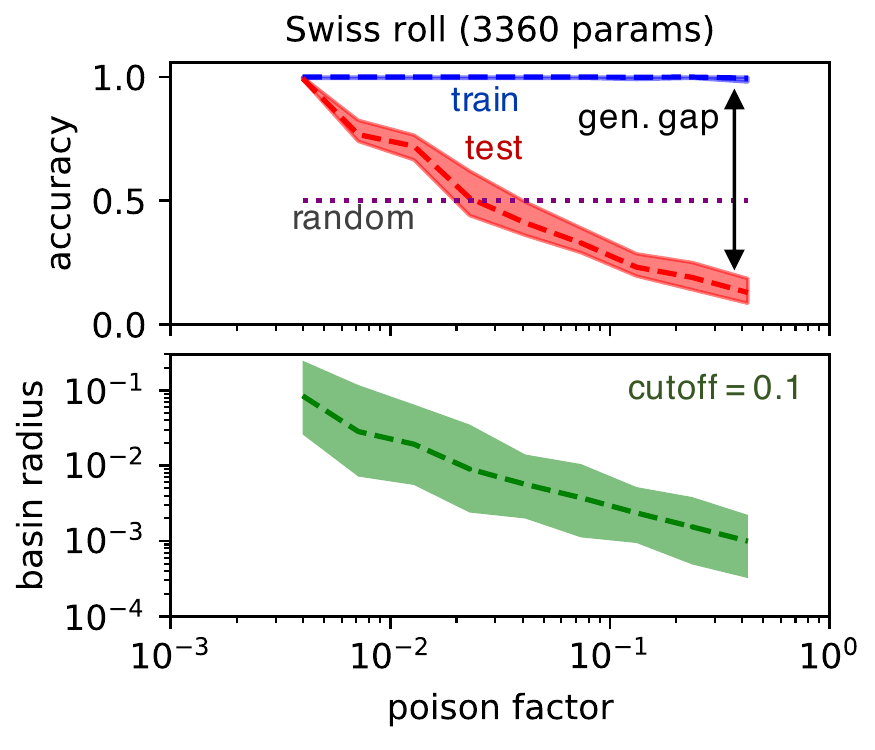}
        \end{subfigure}
        \begin{subfigure}{.48\textwidth}
            \centering
            \includegraphics[width=.99\textwidth]{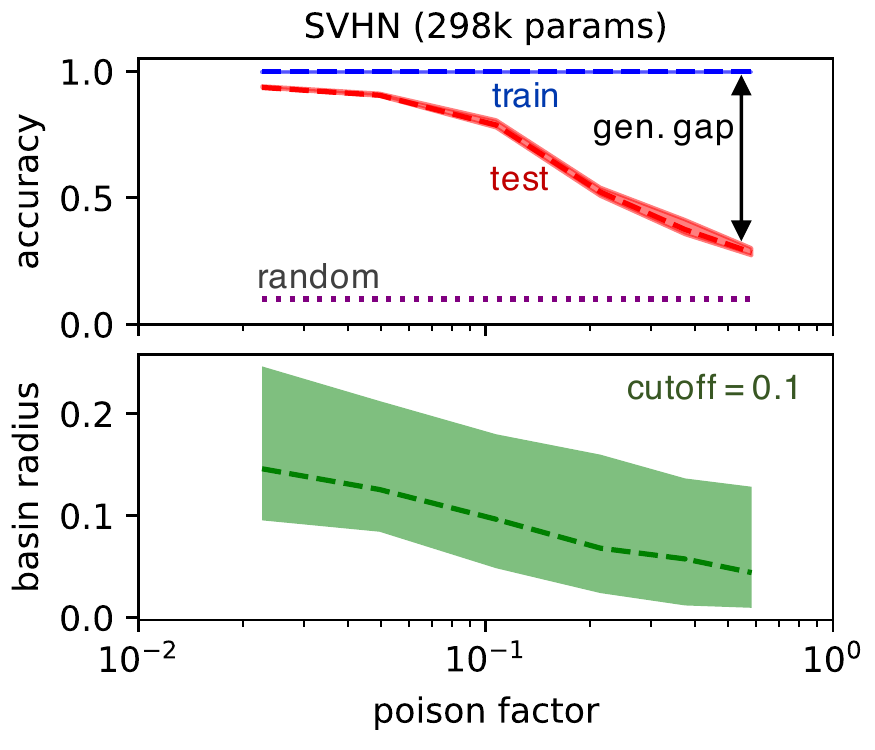}
        \end{subfigure}
        \vskip\baselineskip
        \vspace{-3mm}
        \hspace{-1.5mm}\begin{subfigure}{.49\textwidth}
            \centering
         \includegraphics[width=1.02\textwidth,trim=0mm 0mm 1.0mm 0mm, clip]{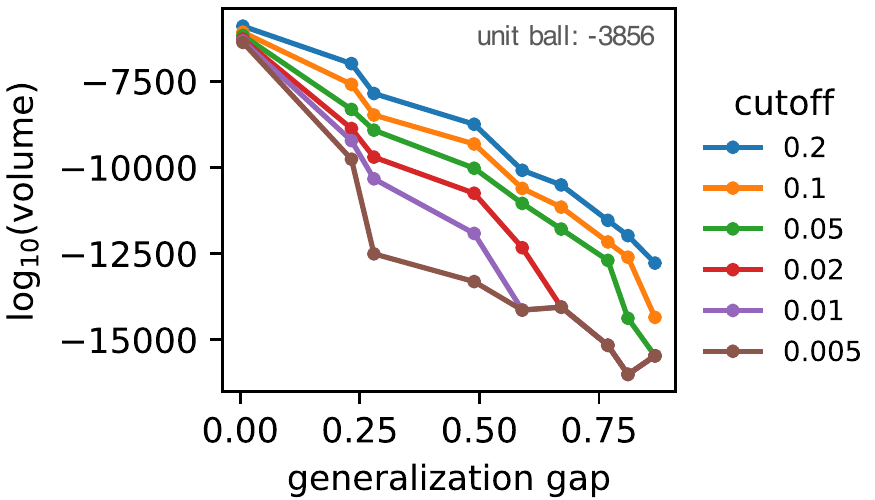}
            \label{fig:surfaceflat-5}
        \end{subfigure}
        \hspace{1mm}
        \begin{subfigure}{.49\textwidth}
            \centering
        \includegraphics[width=1.02\textwidth,trim=0mm 0mm 1mm 0mm, clip]{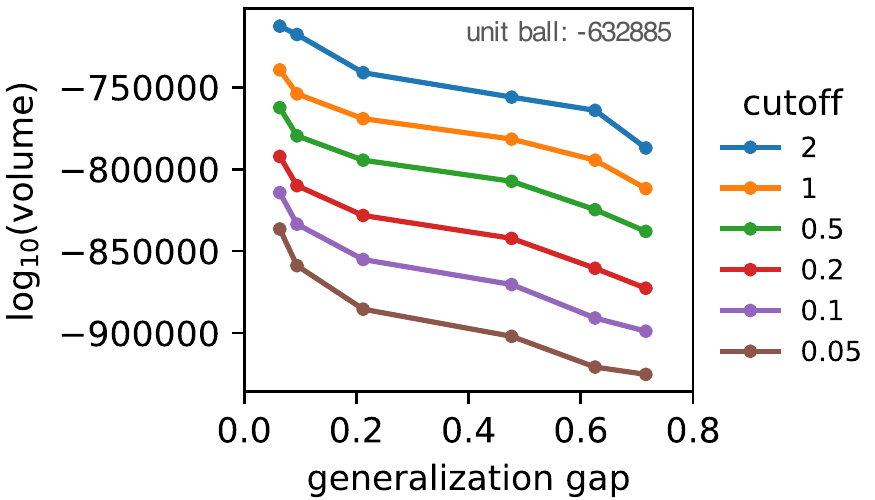}
            \label{fig:surfaceflat-6}
        \end{subfigure}
        \vspace{-4mm}
        \caption{\small Relationship between generalization, sharpness, and volume. Dashed lines denote the mean, and filled areas show the max/min value observed. Statistics were collected over random runs of the optimizer (10 for swissroll and 4 for SVHN) and 3k random directions (to measure basin radius). 
        }
        \label{fig:generalization_and_volume}
        \vspace{-4mm}
    \end{figure}
    
    \begin{figure}[t!]
        \centering
        \includegraphics[width=\textwidth,trim=0mm 7mm 0mm 0mm]{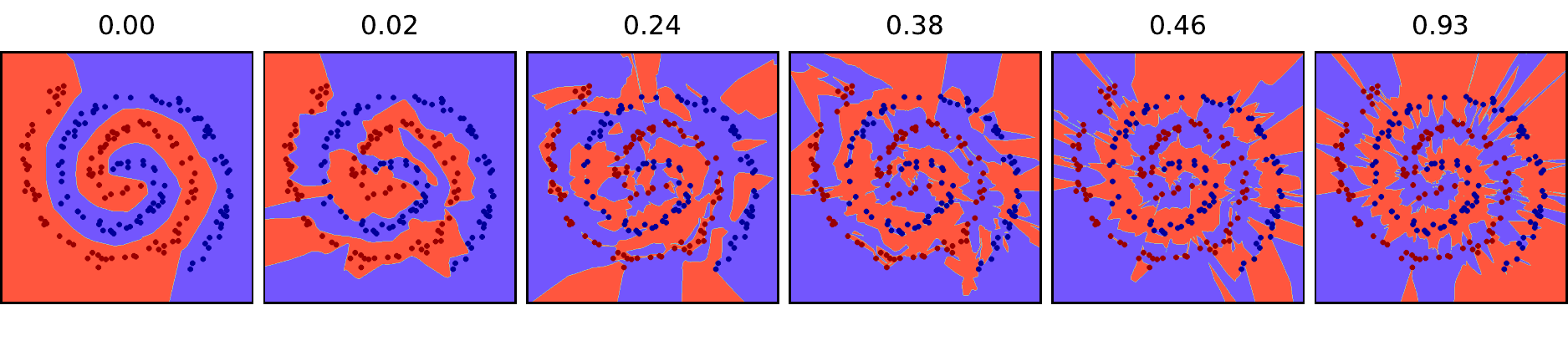}
        \caption{
        \small Swissroll decision boundary for various levels of generalization gap (indicated above plots).}
        \vspace{-4mm}
        \label{fig:swissrollvisual}
    \end{figure}

\section{Implicit regularization and the blessing of dimensionality}
We have seen that neural network loss functions are densely populated with both good and bad minima, and that good minima tend to have ``flat'' loss function geometry.  But what causes optimizers to find these good/flat minima and avoid the bad ones? 

    \begin{wrapfigure}[12]{r}{0.5\textwidth} 
        \centering
            \vspace{-10pt}
            \centering \includegraphics[width=.5\textwidth, trim=1mm 6mm 1mm 0mm, clip]{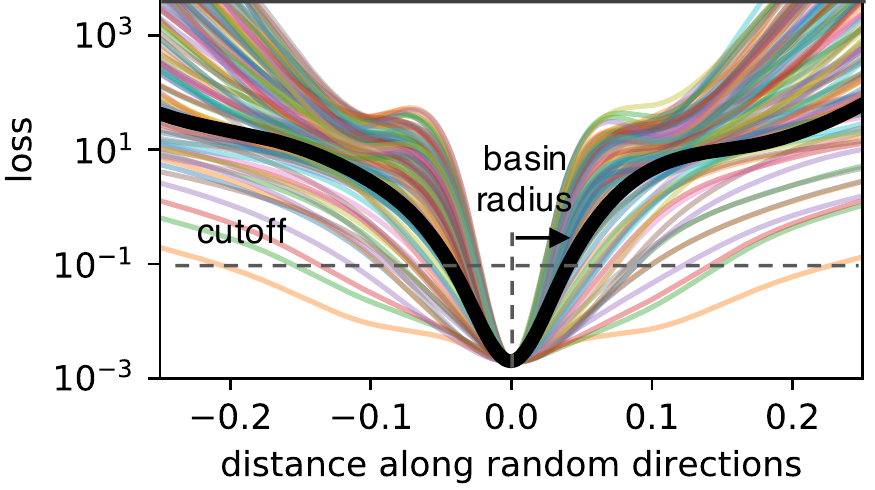}
        \vspace{-10pt}
      \caption{\small SVHN loss along random directions, and the ``basin'' that lies beneath the cutoff loss value.}
        \label{fig:vicinity}
    \end{wrapfigure}
    
The bias of stochastic optimizers towards good minima can be explained by the volume disparity between the basins around good and bad minima.
Flat minima that generalize well lie in wide basins that occupy a large volume of parameter space, while sharp minima lie in narrow basins that occupy a comparatively small volume of parameter space.  As a result, a random search algorithm is more likely to land in the attraction basin for a good minimizer than a bad one.

The volume disparity between good and bad minima is catastrophically magnified by the curse (or, rather, the blessing?) of dimensionality.   The differences in width between good and bad basins does not appear too dramatic in the visualizations in Figures \ref{fig:swissroll-sharpflat} and \ref{fig:sharpsvhn}, or in sharpness visualizations for other datasets (\cite{li2018landscape}).  However, the probability of colliding with a region during a random search does not scale with its width, but rather its \textit{volume}. 
Network parameters live in very high-dimensional spaces where small differences in sharpness between minima translate to exponentially large disparities in the volume of their surrounding basins.   It should be noted that the vanishing probability of finding sets of small width in high dimensions is well studied by probabilists, and is formalized by a variety of  {\em escape theorems} (\cite{gordon1988milman,vershynin2018high}).

To demonstrate the dramatic effect of dimensionality on neural loss landscapes, we quantify the local volume within the low-lying basins surrounding different minima. For this experiment, we define the ``basin'' to be the set of points in a neighborhood
of the minimizer that have loss value below a cutoff of $0.1$ (Fig. \ref{fig:vicinity}). We focus on this definition simply because the volume of this set is efficiently computable.
We calculate the volume of these basins using a Monte-Carlo integration method. Let $r(\phi)$ denote the radius of the basin (distance from minimizer to basin boundary) in the direction of the unit vector $\phi$.  Then the $n$-dimensional volume of the basin is 
$V = \omega_n \mathbb{E}_\phi  [ r^n(\phi)],$
where $\omega_n = \frac{\pi^{n/2}}{\Gamma(1+n/2)}$ is the volume of the unit $n$-ball, and $\Gamma$ is Euler's gamma function.  
We estimate this expectation by calculating $r(\phi)$ for 3k random directions, as illustrated in Figure \ref{fig:vicinity}.


In Figure \ref{fig:generalization_and_volume}, we visualize the combined relationship between generalization and volume for swissroll and SVHN.  By varying the poison factor, we control the test accuracy of each minimizer. As generalization accuracy decreases, we see the radii of the basins decrease as well, indicating that minima become sharper.  Figure \ref{fig:generalization_and_volume} also contains scatter plots showing a severe correlation between generalization and (log) volume for various choices of the basin cutoff value.  For SVHN, the basins surrounding good minima have a volume at least \textit{10,000 orders of magnitude} larger than that of bad minima, rendering it nearly impossible to accidentally stumble upon bad minima.


Finally, we visualize the decision boundaries for several levels of generalization in Figure \ref{fig:swissrollvisual}.  All networks achieve above 99.5\% training accuracy. As the generalization gap increases, the area that belongs to the red class begins encroaching into the area that belongs to the blue class, and vice versa. The margin between the decision boundary and training points also decreases until the training points, though correctly classified, sit on ``islands'' or ``peninsulas'' as discussed above.

 \begin{figure}
        \centering        
        \begin{subfigure}{.32\textwidth}
            \centering
           \includegraphics[width=.99\textwidth, trim=0cm 0cm 0cm 0cm, clip]{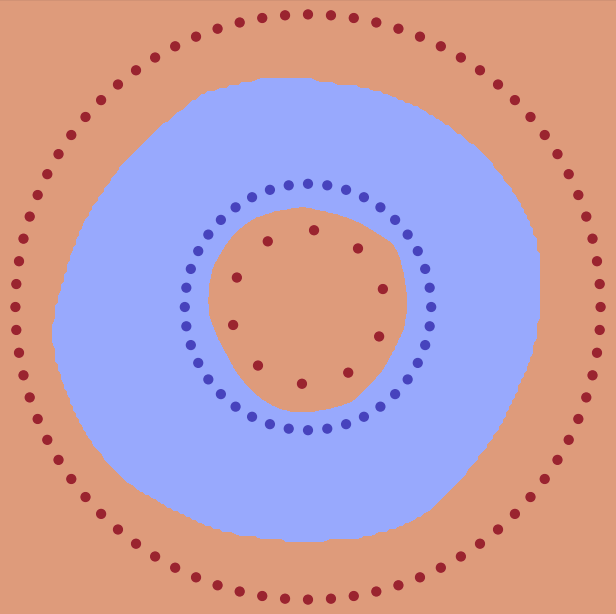}
            \caption{}
            \label{fig:good_circle}
        \end{subfigure}
        \begin{subfigure}{.32\textwidth}
            \centering \includegraphics[width=.99\textwidth, trim=0cm 0cm 0cm 0cm, clip]{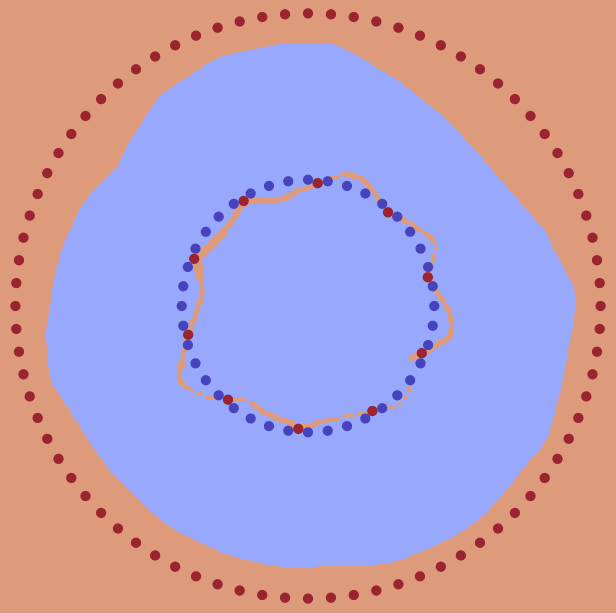}
            \caption{}
            \label{fig:bad_circle}
        \end{subfigure}
      \begin{subfigure}{.32\textwidth}
            \centering \includegraphics[width=.99\textwidth, trim=0cm 0cm 0cm 0cm, clip]{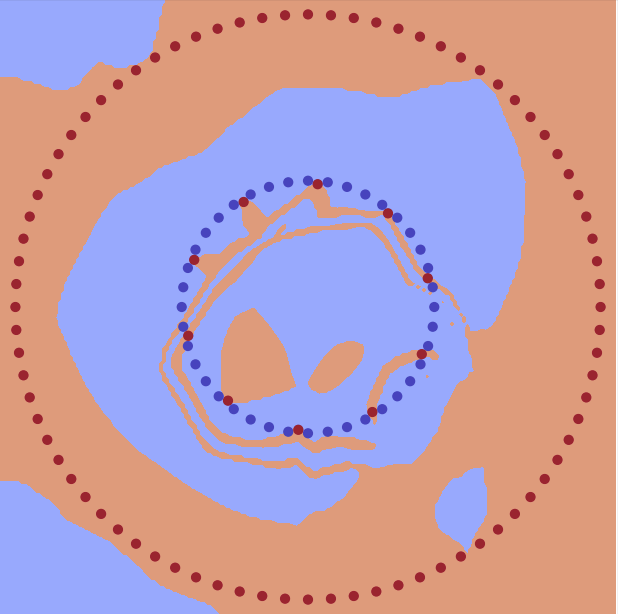}
            \caption{}
            \label{fig:bad_circle-2}
        \end{subfigure}

        \caption{\small 
        A neural network fails to solve a classification problem when the ideal
        solution is ``sharp.'' \vspace{-2mm}
        }
        \label{fig:circs}
\end{figure}
    
\subsection*{A counterfactual experiment:  what can't neural nets solve?}


Neural nets solve complex classification problems by finding ``flat''  minima with class boundaries that assign labels that are stable to parameter perturbations.
Using this intuition, can we formulate a problem that neural nets {\em can't} solve?

Consider the problem of separating the blue and red dots in Figure \ref{fig:circs}. 
When the distance between the inner rings is large, a neural network consistently finds a well-behaved circular boundary as in Fig.\! \ref{fig:circs}a. The wide margin of this classifier makes the minimizer ``flat,'' and the resulting high volume makes it likely to be found by SGD.
We can remove the well-behaved minima from this problem by pinching the margin between the inner red and blue rings.  In this case, two networks trained with random seeds are shown in Figures \ref{fig:circs}b and \ref{fig:circs}c. Now, SGD finds networks that cherry-pick red points, and arc away from the more numerous blue points to maintain a large margin. In contrast, a simple circular decision boundary as in Figure \ref{fig:circs}a would pass extremely close to all points on the inner rings, making such a small margin solution less stable under perturbations and unlikely to be found by SGD.

 \section{Conclusion}
 Using experiments and visualizations, we explored the connection between generalization and loss function geometry.  We also explored a possible explanation for the miraculous generalization behavior of neural nets; high dimensionality crushes bad minima into dust.

While experiments can provide useful insights, they sometimes raise more questions than they answer.  We explored why the ``large margin'' properties of flat minima promote generalization.  But what is the precise metric for ``margin'' that neural networks respect?   Experiments suggest that the small volume of bad minima prevents optimizers from landing in them.  But what is the correct definition of ``volume'' in a space that is invariant to parameter re-scaling and other transforms, and how do we correctly identify the attraction basins for good minima?  Finally and most importantly:  how do we connect all of this back to a rigorous PAC learning framework?

  The goal of this study is to foster appreciation for the complex behaviors of neural networks, and to provide some intuitions for why neural networks generalize. 
 We hope that the experiments contained here will provide inspiration for theoretical progress that leads us to rigorous and definitive answers to the deep questions raised by generalization.

 \section*{Acknowledgements}
 Goldstein and his students were supported by the Office of Naval Research, DARPA's Lifelong Learning Machines and YFA programs, the AFOSR MURI program, and the Sloan Foundation. LF and MG were supported in part by LTS through Maryland Procurement Office and by the NSF DMS 1738003 grant. Studer was supported in part by Xilinx, Inc. and by the US National Science Foundation (NSF) under grants ECCS-1408006, CCF-1535897, CCF-1652065, CNS-1717559, and ECCS-1824379. This work utilized the computational resources of the NIH HPC Biowulf cluster. (\url{http://hpc.nih.gov}). Software from \url{https://comet.ml} and \url{https://sigopt.com} accelerated this work.
    \bibliography{neurips_2019}

\begin{thebibliography}{37}
\providecommand{\natexlab}[1]{#1}
\providecommand{\url}[1]{\texttt{#1}}
\expandafter\ifx\csname urlstyle\endcsname\relax
  \providecommand{\doi}[1]{doi: #1}\else
  \providecommand{\doi}{doi: \begingroup \urlstyle{rm}\Url}\fi

\bibitem[Cybenko(1989)]{cybenko1989universal}
G.~Cybenko.
\newblock Approximation by superpositions of a sigmoidal function.
\newblock \emph{Mathematics of Control, Signals and Systems}, 2\penalty0
  (4):\penalty0 303--314, Dec 1989.

\bibitem[Zhang et~al.(2016)Zhang, Bengio, Hardt, Recht, and
  Vinyals]{zhang2017understanding}
Chiyuan Zhang, Samy Bengio, Moritz Hardt, Benjamin Recht, and Oriol Vinyals.
\newblock Understanding deep learning requires rethinking generalization.
\newblock \emph{International Conference on Learning Representations}, 2016.

\bibitem[Laurent and Brecht(2018)]{laurent2018deep}
Thomas Laurent and James Brecht.
\newblock Deep linear networks with arbitrary loss: All local minima are
  global.
\newblock In \emph{International Conference on Machine Learning}, 2018.

\bibitem[Kawaguchi(2016)]{kawaguchi2016deep}
Kenji Kawaguchi.
\newblock Deep learning without poor local minima.
\newblock In \emph{Advances in neural information processing systems}, pages
  586--594, 2016.

\bibitem[Steinhardt et~al.(2017)Steinhardt, Koh, and
  Liang]{steinhardt2017poison}
Jacob Steinhardt, Pang Wei~W Koh, and Percy~S Liang.
\newblock Certified defenses for data poisoning attacks.
\newblock In \emph{Advances in neural information processing systems}, 2017.

\bibitem[Zhu et~al.(2019)Zhu, Huang, Li, Taylor, Studer, and
  Goldstein]{zhu2019transferable}
Chen Zhu, W~Ronny Huang, Hengduo Li, Gavin Taylor, Christoph Studer, and Tom
  Goldstein.
\newblock Transferable clean-label poisoning attacks on deep neural nets.
\newblock In \emph{International Conference on Machine Learning}, pages
  7614--7623, 2019.

\bibitem[van~der Maaten and Hinton(2008)]{maaten2008tsne}
Laurens van~der Maaten and Geoffrey~E. Hinton.
\newblock Visualizing data using {t-SNE}.
\newblock Proceedings of Machine Learning Research, 2008.

\bibitem[Morcos et~al.(2018)Morcos, Barrett, Rabinowitz, and
  Botvinick]{Morcos2018}
Ari~S. Morcos, David~G.T. Barrett, Neil~C. Rabinowitz, and Matthew Botvinick.
\newblock On the importance of single directions for generalization.
\newblock In \emph{International Conference on Learning Representations}, 2018.

\bibitem[Sun et~al.(2016)Sun, Chen, Wang, Liu, and Liu]{Sun2016}
Shizhao Sun, Wei Chen, Liwei Wang, Xiaoguang Liu, and Tie-Yan Liu.
\newblock On the depth of deep neural networks: A theoretical view.
\newblock In \emph{AAAI}, 2016.

\bibitem[Neyshabur et~al.(2015)Neyshabur, Tomioka, and Srebro]{Neyshabur2015}
Behnam Neyshabur, Ryota Tomioka, and Nathan Srebro.
\newblock Norm-based capacity control in neural networks.
\newblock In \emph{Proceedings of The 28th Conference on Learning Theory},
  2015.

\bibitem[Xie et~al.(2015)Xie, Deng, and Xing]{Xie2015}
Pengtao Xie, Yuntian Deng, and Eric Xing.
\newblock On the generalization error bounds of neural networks under
  diversity-inducing mutual angular regularization.
\newblock \emph{arXiv preprint arXiv:1511.07110}, 2015.

\bibitem[Shalev-Shwartz and Ben-David(2014)]{Shwartz2014}
Shai Shalev-Shwartz and Shai Ben-David.
\newblock \emph{Understanding Machine Learning: From Theory to Algorithms}.
\newblock Cambridge University Press, New York, NY, USA, 2014.

\bibitem[Bartlett et~al.(1998)Bartlett, Maiorov, and Meir]{Bartlett1998}
Peter~L. Bartlett, Vitaly Maiorov, and Ron Meir.
\newblock Almost linear vc dimension bounds for piecewise polynomial networks.
\newblock In \emph{Advances in Neural Information Processing Systems}, 1998.

\bibitem[Harvey et~al.(2017)Harvey, Liaw, and Mehrabian]{Harvey2017}
Nick Harvey, Christopher Liaw, and Abbas Mehrabian.
\newblock Nearly-tight {VC}-dimension bounds for piecewise linear neural
  networks.
\newblock In \emph{Proceedings of the 2017 Conference on Learning Theory},
  2017.

\bibitem[Neyshabur et~al.(2018)Neyshabur, Bhojanapalli, and
  Srebro]{Neyshabur2018}
Behnam Neyshabur, Srinadh Bhojanapalli, and Nathan Srebro.
\newblock A {PAC}-bayesian approach to spectrally-normalized margin bounds for
  neural networks.
\newblock In \emph{International Conference on Learning Representations}, 2018.

\bibitem[Bartlett et~al.(2017)Bartlett, Foster, and Telgarsky]{Bartlett2017}
Peter~L Bartlett, Dylan~J Foster, and Matus~J Telgarsky.
\newblock Spectrally-normalized margin bounds for neural networks.
\newblock In \emph{Advances in Neural Information Processing Systems}. 2017.

\bibitem[Bartlett and Mendelson(2003)]{Bartlett2003}
Peter~L. Bartlett and Shahar Mendelson.
\newblock Rademacher and gaussian complexities: Risk bounds and structural
  results.
\newblock \emph{J. Mach. Learn. Res.}, 3:\penalty0 463--482, March 2003.

\bibitem[Hardt et~al.(2016)Hardt, Recht, and Singer]{Hardt2016}
Moritz Hardt, Benjamin Recht, and Yoram Singer.
\newblock Train faster, generalize better: Stability of stochastic gradient
  descent.
\newblock In \emph{International Conference on International Conference on
  Machine Learning}, 2016.

\bibitem[Kuzborskij and Lampert(2018)]{Kuzborskij2018}
Ilja Kuzborskij and Christoph~H. Lampert.
\newblock Data-dependent stability of stochastic gradient descent.
\newblock In \emph{International Conference on International Conference on
  Machine Learning}, 2018.

\bibitem[Gonen and Shalev-Shwartz(2017)]{Gonen2017}
Alon Gonen and Shai Shalev-Shwartz.
\newblock Fast rates for empirical risk minimization of strict saddle problems.
\newblock In \emph{COLT}, 2017.

\bibitem[Sokolic et~al.(2017)Sokolic, Giryes, Sapiro, and
  Rodrigues]{Sokolic2016}
Jure Sokolic, Raja Giryes, Guillermo Sapiro, and Miguel Rodrigues.
\newblock Generalization error of invariant classifiers.
\newblock In \emph{Artificial Intelligence and Statistics}, pages 1094--1103,
  2017.

\bibitem[McAllester(1998)]{McAllester1998}
David~A. McAllester.
\newblock Some pac-bayesian theorems.
\newblock In \emph{Proceedings of the Eleventh Annual Conference on
  Computational Learning Theory}, COLT' 98, pages 230--234, 1998.

\bibitem[McAllester(1999)]{McAllester1999}
David~A. McAllester.
\newblock Pac-bayesian model averaging.
\newblock In \emph{Proceedings of the Twelfth Annual Conference on
  Computational Learning Theory}, COLT '99, pages 164--170, New York, NY, USA,
  1999.

\bibitem[Neyshabur et~al.(2017)Neyshabur, Bhojanapalli, McAllester, and
  Srebro]{neyshabur2017exploring}
Behnam Neyshabur, Srinadh Bhojanapalli, David McAllester, and Nati Srebro.
\newblock Exploring generalization in deep learning.
\newblock In \emph{Advances in Neural Information Processing Systems}, pages
  5947--5956, 2017.

\bibitem[Golowich et~al.(2018)Golowich, Rakhlin, and Shamir]{golowich2018}
Noah Golowich, Alexander Rakhlin, and Ohad Shamir.
\newblock Size-independent sample complexity of neural networks.
\newblock In \emph{Proceedings of the 31st Conference On Learning Theory},
  2018.

\bibitem[Dziugaite and Roy(2017)]{dziugaite2017computing}
Gintare~Karolina Dziugaite and Daniel~M Roy.
\newblock Computing nonvacuous generalization bounds for deep (stochastic)
  neural networks with many more parameters than training data.
\newblock \emph{Conference on Uncertainty in Artificial Intelligence (UAI)},
  2017.

\bibitem[Arora et~al.(2018)Arora, Ge, Neyshabur, and Zhang]{Arora2018}
Sanjeev Arora, Rong Ge, Behnam Neyshabur, and Yi~Zhang.
\newblock Stronger generalization bounds for deep nets via a compression
  approach.
\newblock In \emph{International Conference on Machine Learning}, pages
  254--263, 2018.

\bibitem[Hochreiter and Schmidhuber(1997)]{hoch1997flat}
Sepp Hochreiter and J{\"u}rgen Schmidhuber.
\newblock Flat minima.
\newblock \emph{Neural Computation}, 9:\penalty0 1--42, 1997.

\bibitem[Keskar et~al.(2017)Keskar, Mudigere, Nocedal, Smelyanskiy, and
  Tang]{keskar2016largebatch}
Nitish~Shirish Keskar, Dheevatsa Mudigere, Jorge Nocedal, Mikhail Smelyanskiy,
  and Ping Tak~Peter Tang.
\newblock On large-batch training for deep learning: Generalization gap and
  sharp minima.
\newblock \emph{International Conference on Learning Representations}, 2017.

\bibitem[Izmailov et~al.(2018)Izmailov, Podoprikhin, Garipov, Vetrov, and
  Wilson]{izmailov2018swa}
Pavel Izmailov, Dmitrii Podoprikhin, Timur Garipov, Dmitry Vetrov, and
  Andrew~Gordon Wilson.
\newblock Averaging weights leads to wider optima and better generalization.
\newblock \emph{arXiv preprint arXiv:1803.05407}, 2018.

\bibitem[Wang et~al.(2018)Wang, Keskar, Xiong, and Socher]{wang2018identifying}
Huan Wang, Nitish~Shirish Keskar, Caiming Xiong, and Richard Socher.
\newblock Identifying generalization properties in neural networks.
\newblock \emph{arXiv preprint arXiv:1809.07402}, 2018.

\bibitem[Li et~al.(2018)Li, Xu, Taylor, Studer, and Goldstein]{li2018landscape}
Hao Li, Zheng Xu, Gavin Taylor, Christoph Studer, and Tom Goldstein.
\newblock Visualizing the loss landscape of neural nets.
\newblock In \emph{Advances in Neural Information Processing Systems}, pages
  6389--6399. 2018.

\bibitem[Chaudhari et~al.(2017)Chaudhari, Choromanska, Soatto, LeCun, Baldassi,
  Borgs, Chayes, Sagun, and Zecchina]{chaudhari2017entropy}
P~Chaudhari, Anna Choromanska, S~Soatto, Yann LeCun, C~Baldassi, C~Borgs,
  J~Chayes, Levent Sagun, and R~Zecchina.
\newblock Entropy-sgd: Biasing gradient descent into wide valleys.
\newblock In \emph{International Conference on Learning Representations
  (ICLR)}, 2017.

\bibitem[Dinh et~al.(2017)Dinh, Pascanu, Bengio, and Bengio]{dinh2017sharp}
Laurent Dinh, Razvan Pascanu, Samy Bengio, and Yoshua Bengio.
\newblock Sharp minima can generalize for deep nets.
\newblock In \emph{International Conference on Machine Learning}, 2017.

\bibitem[Netzer et~al.(2011)Netzer, Wang, Coates, Bissacco, Wu, and
  Ng]{netzer2011svhn}
Yuval Netzer, Tao Wang, Adam Coates, Alessandro Bissacco, Bo~Wu, and Andrew~Y.
  Ng.
\newblock Reading digits in natural images with unsupervised feature learning.
\newblock NIPS Workshop on Deep Learning and Unsupervised Feature Learning
  2011, 2011.

\bibitem[Gordon(1988)]{gordon1988milman}
Yehoram Gordon.
\newblock On {Milman's} inequality and random subspaces which escape through a
  mesh in $\mathbb{R}$ n.
\newblock In \emph{Geometric Aspects of Functional Analysis}, pages 84--106.
  Springer, 1988.

\bibitem[Vershynin(2018)]{vershynin2018high}
Roman Vershynin.
\newblock \emph{High-dimensional probability: An introduction with applications
  in data science}, volume~47.
\newblock Cambridge University Press, 2018.

\end{thebibliography}


\end{document}